# Superimposition of eye fundus images for longitudinal analysis from large public health databases

G Noyel[1, 4], R Thomas[2], G Bhakta[3], A Crowder[3], D Owens[2] and P Boyle[1, 4]

[1]International Prevention Research Institute, 95 Cours Lafayette, Lyon, 69006 France

[2]Diabetes Research Group, Institute of Life Sciences, College of Medicine, Swansea University, SA2 8PP, Wales, UK

[3]Diabetic Eye Screening Wales, 1 Fairway Court, Tonteg Road, Upper Boat, Treforest, Pontypridd CF37 5UA, Wales, UK

[4]Strathclyde Institute of Global Public Health, University of Strathclyde, Glasgow, UK

e-mail: guillaume.noyel@i-pri.org, r.l.thomas@swansea.ac.uk, owensDR@cardiff.ac.uk, gavin.bhakta@drssw.wales.nhs.uk, andrew.crowder@drssw.wales.nhs.uk, peter.boyle@i-pri.org

**Abstract**

In this paper, a method for superimposition (i.e. registration) of eye fundus images from persons with diabetes screened over many years for Diabetic Retinopathy is presented. The method is fully automatic and robust to camera changes and colour variations across the images both in space and time. All the stages of the process are designed for longitudinal analysis of cohort public health databases where retinal examinations are made approximately at a year of interval. The method relies on a model correcting two radial distortions and an affine transformation between pairs of images which is robustly fitted on salient points. Each stage involves linear estimators followed by non-linear optimisation. The model of image warping is also invertible for fast computation. The method has been validated 1) on a simulated montage and 2) on public health databases with 69 patients with high quality images (with 271 pairs acquired mostly with different type of cameras and with 268 pairs acquired mostly with the same type of cameras) with a success rates of 92 % and 98 % and 5 patients (with 20 pairs) with low quality images with a success rate of 100%. Compared to two methods, ours gives better results.

Superimposition of Eye Fundus Images for Longitudinal Analysis

## 1 Introduction

Diabetic Retinopathy (DR) as one of the major causes of visual impairment in the world represents a major public health challenge. It is a complication of both types of diabetes mellitus, which affects the light perception part of the eye (retina). DR may lead to the development of sight threatening lesions and without adequate and timely treatment the patients could lose their sight and eventually become blind (International Diabetes Federation and The Fred Hollows Foundation, 2015; Scanlon *et al.*, 2009). DR is often asymptomatic until an advanced stage, thereby screening to detect sight threatening DR at an early stage is essential which has resulted in the introduction of DR Screening services in many countries such as UK (Harding *et al.*, 2003), USA, the Netherlands, France, etc. The commonest screening method involves acquiring eye fundus images on an annual or biennial basis.

As these DR screening programs have been in existence over several years, performing longitudinal analysis of the eye fundus images of the same patient is now possible. However, in order to accurately compare the evolution of DR over time, the images must be perfectly super-imposed.

The direct superimposition of two images of the same patient never gives good results (see Figure 1). Indeed, for two separate photographic-eye examinations the patient is never in exactly in same position and also the camera may differ. Therefore, the super-imposition method has to take into account the different causes of the deformation such as: i) the position of the patient: by taking into account rotation, translation and scaling, ii) the change of the camera: by using scaling and radial correction, iii) the projection of a 3D scene assimilated to a sphere (the retina of the eye) onto the 2D plane of the sensor of the camera: by using radial correction, iv) the radial deformation due to the optics of the camera: by using radial correction and v) the colour variability between images due to the light intensity and sensor.

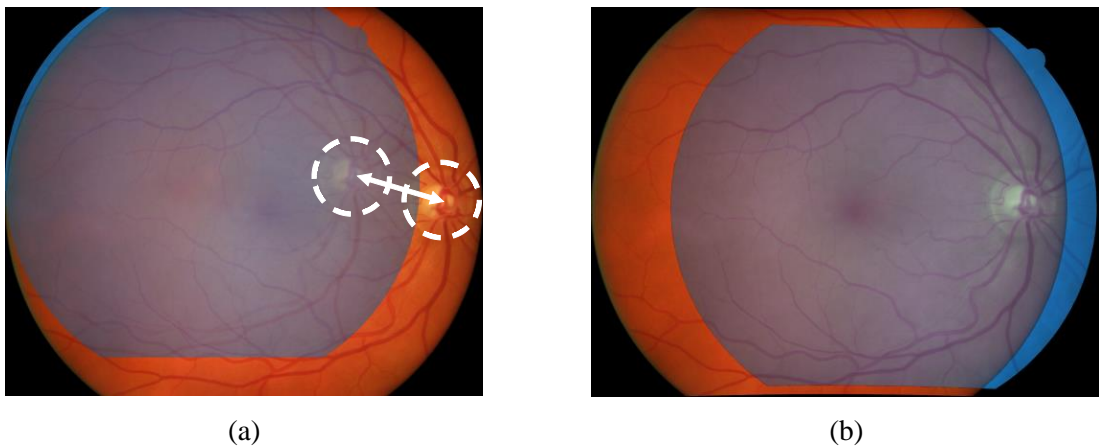

(a)                                                        (b)

Figure 1. Superimposition of eye fundus images acquired with different cameras and image resolution. (a) Naive superimposition of images. (b) Perfect superimposition of images by using a model. For display purpose, in each subfigure (a) and (b), one of the image of the pair is in false colours (i.e. bluish colours).

To perform a superimposition – also named registration – two stages are regarded as necessary: a model of deformation and a matching criterion to fit the model. Registration of medical images has been a very active field for research during the last decades especially in the field of radiology. For an overview of medical image registration methods see e.g. (Maintz and Viergever, 1998; Oliveira and Tavares, 2014; Viergever *et al.*, 2016). (Pluim *et al.*, 2003) have also written a survey about mutual-information-based registration of medical images and (Sotiras *et al.*, 2013) about deformable registration. In this paper, we will focus on the registration of eye fundus images.

Several models of super-imposition between pairs of eye fundus images are in existence. The earliest methods relied on fluorescein images and based on a composition of translation, rotation and scaling - i.e. an affine transformation model (Zana and Klein, 1999a, b). The bifurcations of the vessels were used to match the points and fit the model. Another matching criterion consists in the minimisation of image intensity differences (Cideciyan, 1995; Matsopoulos *et al.*, 1999; Ritter *et al.*, 1999; Adal *et al.*, 2014). Matching by salient points using SURF - Speeded Up Robust Features - detector, was used in (Cattin *et al.*, 2006).

Other methods are based on a similarity (i.e. a rotation and a translation) (Matsopoulos *et al.*, 1999; Lloret *et al.*, 2000) and an elastic model of deformation (You *et al.*, 2005; Fang and Tang, 2006). In Jian *et al.* (2010); Ghassabi *et al.* (2013), new descriptors PIIFD (Partial Intensity Invariant Feature Descriptor) have been introduced for multimodal image superimposition between auto-fluorescence, infrared and red-free images (i.e. the green component of a colour eye fundus images) using an affine and a quadratic model of deformation. This method, powerful for multimodal image registration, has not been designed and tested for the superimposition of colour eye fundus images.

More recently, it has been shown that a quadratic model gives better registration results (Adal *et al.*, 2014; Can *et al.*, 2002; Stewart *et al.*, 2003; Chanwimaluang *et al.*, 2006; Ryan *et al.*, 2004). The difficulty inherent in these models is to estimate their parameters. To overcome such a limitation, a radial distortion model has been introduced by Lee *et al.* (2007) and compared to previous methods in Lee *et al.* (2010). It consists of adding a radial model to the affine transformation in order to correct the effects of radial distortion due to the geometry of the camera and of the eye. Superimposition of eye fundus images have also been performed in the tri-dimensional space using a model of near planar surface (Yuping and Medioni, 2008) or an ellipsoid model (Hernandez-Matas *et al.*, 2016).

However, for all these methods the superimposition of eye fundus images is performed with colour images acquired by the same camera during the same examination. Superimposing colour images acquired by different cameras with at least a year of interval and stored in large screening databases still remains a challenging problem. Two new issues are appearing: i) the radial distortion must be corrected for the two different lenses, because the camera has changed and ii) the colour of the eye may have changed due to a different source of light and to an evolution of the anatomy of the eye (e.g. cataract removal, evolution of the retinopathy to a more severe stage, etc.)

Currently, for colour eye fundus images, no existing method has been designed to address the problem of the change of camera lenses, camera and light for retinal examination used for DR screening (i.e. at a year of interval). In addition, in the existing methods, even when one radial distortion is corrected by a radial distortion correction or by a second order model, the registration criterion requires a similar and uniform intensity (i.e. colour) of the images, particularly if the extraction of anatomical features such as the vessels is needed. These methods are working in laboratory conditions (several acquisitions on the same day with the same equipment) but failed when they are used on public health databases constituted by routine screening across years of patients with Diabetes.

In this paper, our **contribution** has been to address this challenge by presenting a robust superimposition method designed for longitudinal screening of large public health image based databases. After having taken into account the variation of colour, we introduce a new model of image registration based on two radial distortions and a homography. Importantly, this model is invertible which is useful for image deformation (i.e. image warping) and makes it suitable for fast computation, a key factor for an efficient analysis of large databases. We also introduce linear equations to estimate the parameters of the model in a fast and simple way. Finally, we provide a complete framework for robust image superimposition. The registration criterion is based on a standard salient point extraction, which does not rely on the extraction of anatomical features.

The paper is organised as follows: after presenting our method to superimpose pairs of images, it will be validated on a simulated montage and using different patient databases. It will be compared with two well-established methods of image registration. Then, the results will be interpreted and discussed before the conclusion.

## 2 Methods

During a photographic eye examination, eye fundus (retina) two images of both eyes are acquired in two positions i.e. a 45 degree "nasal" and "macular" field (Figure 2). As patients with diabetes are annually screened this make series of images available for longitudinal analysis. This is a different problem from large mosaics of eye fundus images acquired during the same examination with the same camera (Chanwimaluang *et al.*, 2006). Consequently, the aim of this study was to develop a superimposition method of images in the same positions while being captured during two different exams and often with different cameras and resolutions.

A schematic description of the study is represented in Figure 3. The different stages have been designed to provide efficient solutions to the superimposition of images acquired for practical screening. Between two examinations, the camera and lighting might have been changed producing differences in colour requiring a pre-processing stage consisting of (1) normalizing the colour of the eye fundus image (Noyel *et al.*, 2015). In addition, the position, resolution and radial distortion between images might be different requiring an image transformation estimated thanks to the (2) extraction of characteristic points in pairs of images (3) a matching procedure, (4) the use of a model correcting radial distortion of both images and (5) the estimation of the parameters of the model by a robust optimisation. Eventually, the method is validated (6) on a simulated montage and the superimpositions of the image of the database are verified (7) and compared to state of the art methods.

Superimposition of Eye Fundus Images for Longitudinal Analysis

In this paper, the image pairs of the same eye in the same position, "nasal" or "macular", are registered independently.

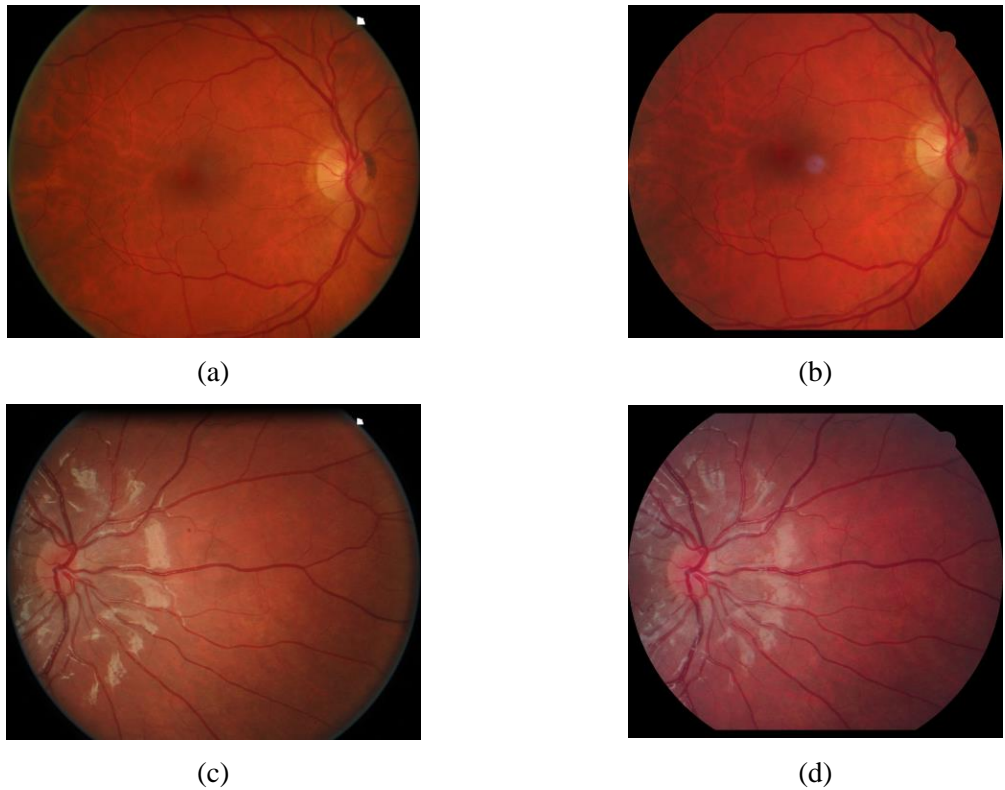

<div align="center">(a)              (b)</div>

<div align="center">(c)              (d)</div>

Figure 2. (a) and (b) macular eye fundus images acquired during two exams with 1 year of interval on a patient. (c) and (d) nasal eye fundus images acquired during two exams with 1 year of interval on another patient. For both patients, different cameras, resolution and lighting conditions were used.

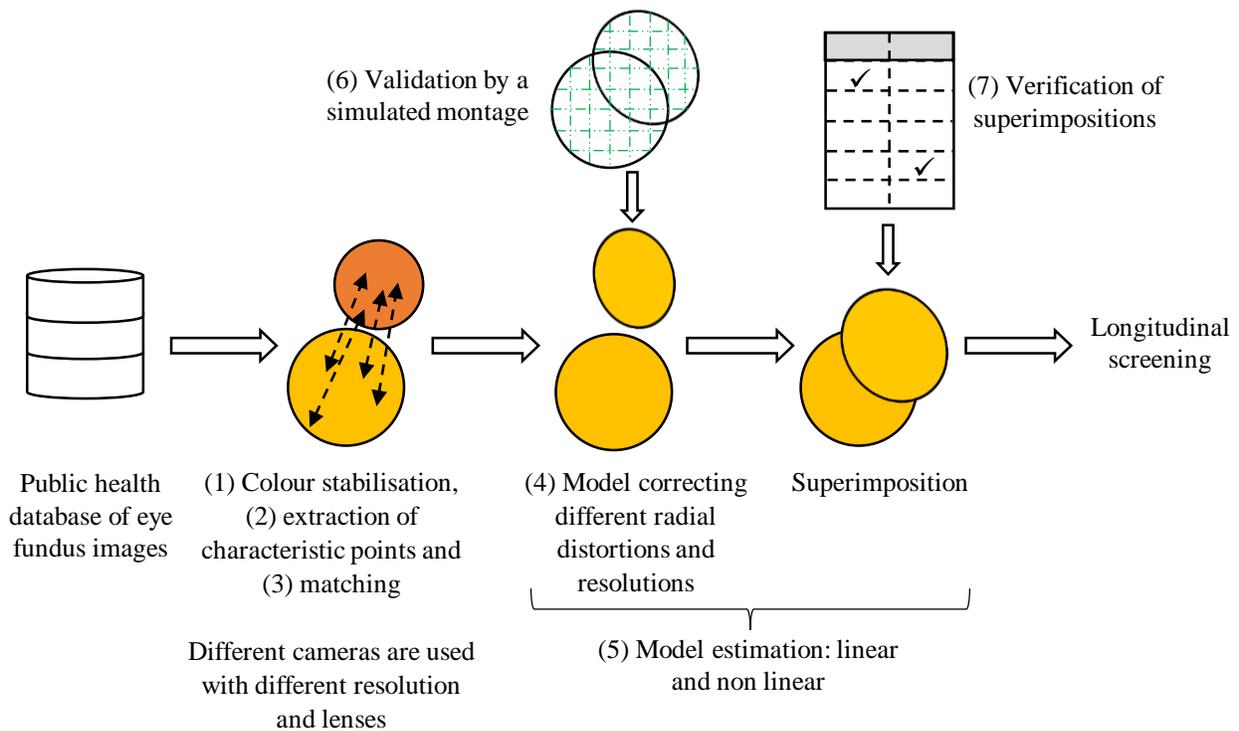

Figure 3. Framework of eye fundus superimposition of images for longitudinal screening. After colour stabilisation (pre-processing) and extraction of characteristic points, the points are matched and the model is estimated. The differences are corrected in term of radial distortions and resolutions between the images.

Superimposition of Eye Fundus Images for Longitudinal Analysis

## 2.1 Pre-processing

The brightness of eye fundus images is non-uniform due to various reasons: disease such as cataract, motion of the patient, acquisition conditions and differences in absorptions of the light in the eye (Walter, 2003; Walter and Klein, 2005). Some parts of the images appear as bright while others are dark. Moreover, the possible change of the eye fundus camera between two separate examinations may contribute to a change in the colour between two images of the same eye (Figure 4). We have used a method (Noyel *et al.*, 2015) to correct the variations of colour contrast between the images. The method is based on a colour model consistent with the physical principles of image formation. The contrast of the dark or the bright elements are adjusted in a way that provides a similar colour aspect to lesions such as micro-aneurysms or to anatomical structures such as vessels or veins. Results can be seen in Figure 4. Our colour methods goes further than previous existing grey-level methods using polynomial adjustment (Walter *et al.*, 2007), mathematical morphology (Zhang *et al.*, 2014) or histogram equalisation (Zuiderveld, 1994).

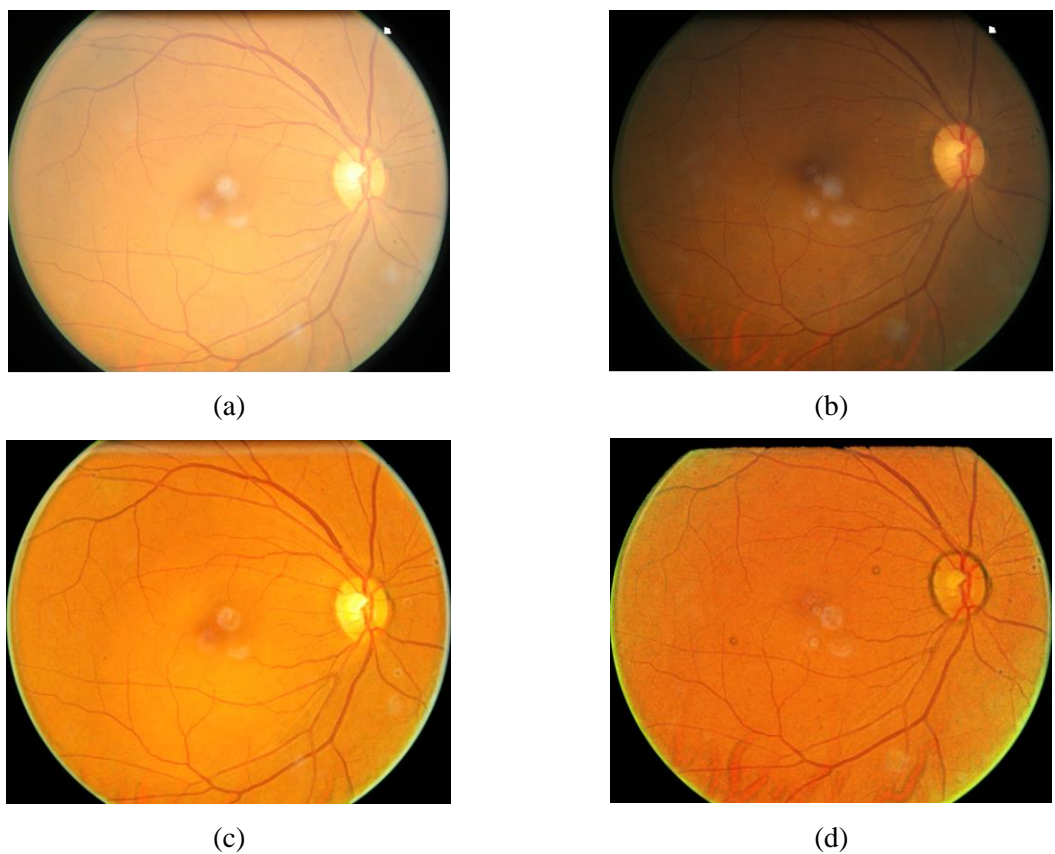

(a)        (b)

(c)        (d)

Figure 4. (a), (b) Eye fundus images of the same patient acquired during two exams with 1.5 years of interval with the same camera. These low quality images have been acquired in harsh conditions. (c), (d) Colour stabilisation of eye fundus images (a) and (b).

## 2.2 Extraction of characteristic points

The pre-processing is followed by the extraction of several salient points (Figure 5) using the Scale-Invariant Feature Transform - SIFT - algorithm (Lowe, 2004; Vedaldi and Fulkerson, 2008). The SIFT algorithm has been designed to be robust to the variation of observation angle and to some variations in lighting. Briefly SIFT consists of extracting key points based on a multiscale analysis. Then, series of descriptors are computed for each salient point. These descriptors are used for point matching.

## 2.3 Point matching with a preliminary estimate of the image deformation

Point matching is a necessary step before the estimation of the model of deformation. With the SIFT method to extract characteristic points, (Lowe, 2004) has proposed a matching method. However, it was not robust enough to estimate the model of deformation in the database. Therefore, a three-step procedure of matching has been proposed:

(a) A first matching by Lowe's method followed by a refined selection of the correspondence vectors according to their size and orientation (see hereinafter).

(b) A preliminary estimate of the image deformation by a homography. The position of the characteristic points is transformed using a homography (see hereinafter).

(c) Step (a) is run a second time using the transformed points.

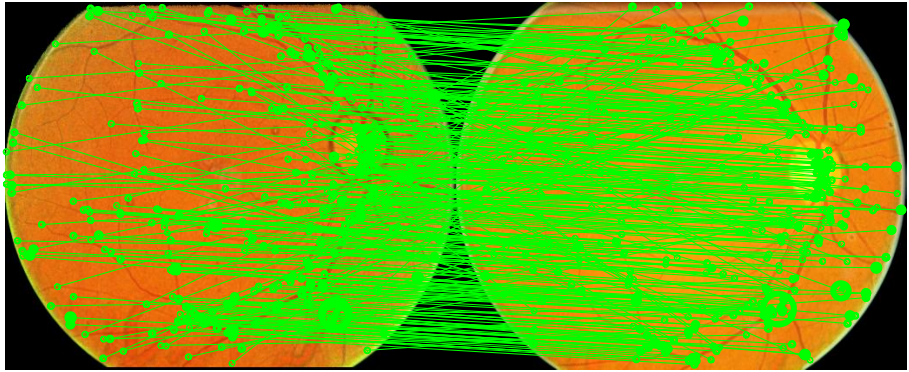

(a)

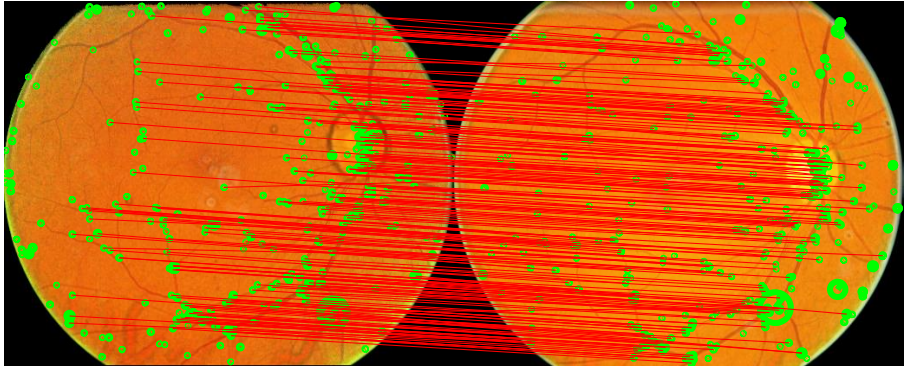

(b)

Figure 5. Point extraction and matching between images. (a) Initial matching (green arrows). (b) Matching after simplification (red arrows).

### 2.3.1 Selection of the correspondence vectors

First of all, the region of interest of both images are extracted and joint together after resizing and padding to have the same vertical size (Figure 5). As the image deformation is closed to a similarity (i.e. rotation, translation and scaling), the length of ratios and angles are invariant by similarity (Hartley and Zisserman, 2004). This property is used to remove the correspondence vectors $v$ between inconsistent matched points. With the two images joint together, a two-step selection is performed on the lengths $l$ and orientations $\theta$ of the correspondence vectors:

(i) only the vectors $v$ whose length $l_v$ and orientation $\theta_v$ is in the interval $\{|l_v - E\{l_v\}| \leq \sigma\{l_v\}$ and $|\theta_v - E\{\theta_v\}| \leq 5°\}$ are kept. $E\{\}$ is the mean and $\sigma\{\}$ the standard deviation of a variable.

(ii) Among the selected vectors $\tilde{v}$, only the vectors whose length $l_{\tilde{v}}$ and orientation $\theta_{\tilde{v}}$ is in the interval $\{|l_{\tilde{v}} - E\{l_{\tilde{v}}\}| \leq \max(3\sigma\{l_{\tilde{v}}\}, 5\% \times ysize)$ and $|\theta_{\tilde{v}} - E\{\theta_{\tilde{v}}\}| \leq \max(5°, \sigma\{\theta_{\tilde{v}}\})\}$ are kept. $ysize$ is the number of lines in the image.

Notice: The value 5° has been empirically selected and perfectly works for all the processed images.

### 2.3.2 Estimation of a homography

The affine homography (or affine transformation) $H$ is defined as:

$$H = \begin{bmatrix} A & T \\ O^T & 1 \end{bmatrix} = \begin{bmatrix} a_{11} & a_{12} & t_x \\ a_{21} & a_{22} & t_y \\ 0 & 0 & 1 \end{bmatrix} \tag{1}$$

$A = \begin{bmatrix} a_{11} & a_{12} \\ a_{21} & a_{22} \end{bmatrix}, \forall i,j \in \{1,2\}, \ a_{ij} \in \mathbb{R}$, is a $2 \times 2$ non-singular matrix (i.e. with a nonzero determinant) and represents the linear applications. $T = \begin{bmatrix} t_x \\ t_y \end{bmatrix}, \ t_x, t_y \in \mathbb{R}$, is a translation.

The estimation of a homography is performed using a robust maximum likelihood estimation with the RANdom SAmple Consensus (RANSAC) algorithm applied on the "linear normalised DLT (Direct Linear Transform)" algorithm (Hartley and Zisserman (2004)). This is the initialisation step of the "gold standard algorithm" of Hartley and Zisserman (2004). This initialisation algorithm estimates around 100 homographies on randomly selected 4 pairs of points. Then, the homography with the minimum error, when transforming the matched points, is selected. This allows a removal of the incorrect matchings. However, as there is also a radial distortion in the image, the deformation is not entirely modelled by a homography. Therefore, some homographies must be discarded. In particular, those with a scaling factor on the $x$ and $y$ axis with a relative difference greater than 1%, because the sensor resolution is almost the same in $x$ and in $y$. For this purpose, several estimates (until 50) using the initialisation algorithm are performed if the relative difference between the scaling factors is greater than 1%. If the value of 1% is never reached, then the homography with the smallest relative difference between the scaling factors is kept. Finally, a non-linear estimation is performed with the algorithm of Levenberg-Marquardt (More, 1977; Bonnans *et al.*, 2006). It serves as a good initialisation of the estimation of the complete deformation model.

*2.4  Model of deformation*

The model of deformation ensures a correct superimposition between the images. Several deformations are taken into account: (i) the difference in terms of positions of the eye between a pair of images will be corrected by an affine transformation (i.e. and homography) and (ii) the radial deformations due to the projection of the eye into the camera and due to the optics of the camera (Hartley and Zisserman, 2004) will be corrected using a radial transformation. For this purpose, Lee *et al.* (2007); (Lee *et al.*, 2008), have proposed a model coupling a unique radial transformation for both images and a homography. Lee *et al.* (2010) have made the comparison with two other second order models.

Their approach was extended by defining a model with one homography $H$ and two radial distortions, i.e. one for each image. Indeed, the camera may have changed between screening exams on a large number of patients.

The radial distortion due the background of the sphere surface of the eye and of the radial distortion of the camera was modelled by a *division model* (Fitzgibbon, 2001) in the following way:

$$\bar{P}^d = \left(1 + k\left(r^d\right)^2\right) . \bar{P}^u \tag{2}$$

$P^d \in \mathbb{R}^2$ are the distorted coordinates in the original (i.e. distorted) image. $P^u \in \mathbb{R}^2$ are the undistorted coordinates in the undistorted image. $\bar{P}^d \in \mathbb{R}^2$ are the distorted coordinates centred on the image centre $C$: $\bar{P}^d = P^d - C$. $\bar{P}^u \in \mathbb{R}^2$ are the undistorted coordinates centred on the image centre $C$: $\bar{P}^u = P^u - C$. $r^d = \|P^d - C\| = \|\bar{P}^d\| \in \mathbb{R}$, is the distance of the deformed coordinates $P^d$ from the optic centre $C$ (i.e. the image centre). $k$ is a real parameter of distortion. Its normalised version, $\tilde{k} = k(1 + \|C\|^2)$, is in the interval $[-0.2 \ ; 0.2]$.

The model was named division model because the distorted coordinates were divided by the radial distortion $\bar{P}^d / \left(1 + k\left(r^d\right)^2\right) = \bar{P}^u$. The distorted image corresponds to the original image and the undistorted image is the image after the correction of radial distortion.

Given $P_1^d$ and $P_2^d$ the coordinates of the points in the original (i.e. deformed) images 1 and 2, $k_1$ and $k_2$ the distortion parameters, $C_1$ and $C_2$ the image centres, the model mapping image 1 into image 2 is defined as follows:

$$\frac{\bar{P}_2^d}{\left(1 + k_2\left(r_2^d\right)^2\right)} + C_2 = H\left[\frac{\bar{P}_1^d}{\left(1 + k_1\left(r_1^d\right)^2\right)} + C_1\right] \tag{3}$$

If the same camera is used to acquire both images, the distortion parameters are equal, $k_1 = k_2 = k$, and the model corresponds to the model of Lee *et al.* (2007). However, in our approach, the model is estimated after having extracted and matched the points in the pair of original images (target and reference). Therefore, the radial distortion correction is performed after the detection of the feature correspondence points in the original (i.e. distorted) image. It has been programmed in the subsequent way.

After point matching, the model is estimated by minimising the deformation error $err$ between the undistorted points in the image 2 estimated by the model, $(\bar{P}_2^u)^{est} = H\left[\frac{\bar{P}_1^d}{\left(1+k_1(r_1^d)^2\right)} + C_1\right] - C_2$, and the undistorted points extracted in the image 2, $\bar{P}_2^u = \bar{P}_2^d / \left(1 + k_2(r_2^d)^2\right)$. The deformation error is $err = \|(\bar{P}_2^u)^{est} - \bar{P}_2^u\|$.

### 2.5 Estimation of the model parameters

The model parameters are estimated in (Figure 6) by a different method of (Lee *et al.*, 2007; Abramoff *et al.*, 2012). In particular, during the initialisation step, the type of the model is selected: either a single distortion model, if the images are acquired by the same camera, or a two distortions model if two cameras are used. In addition, the radial distortion is estimated after the homography without needing any preliminary initialisation by a calibration of the camera (Hartley and Zisserman, 2004). Moreover, we have introduced linear estimators at each step of the process.

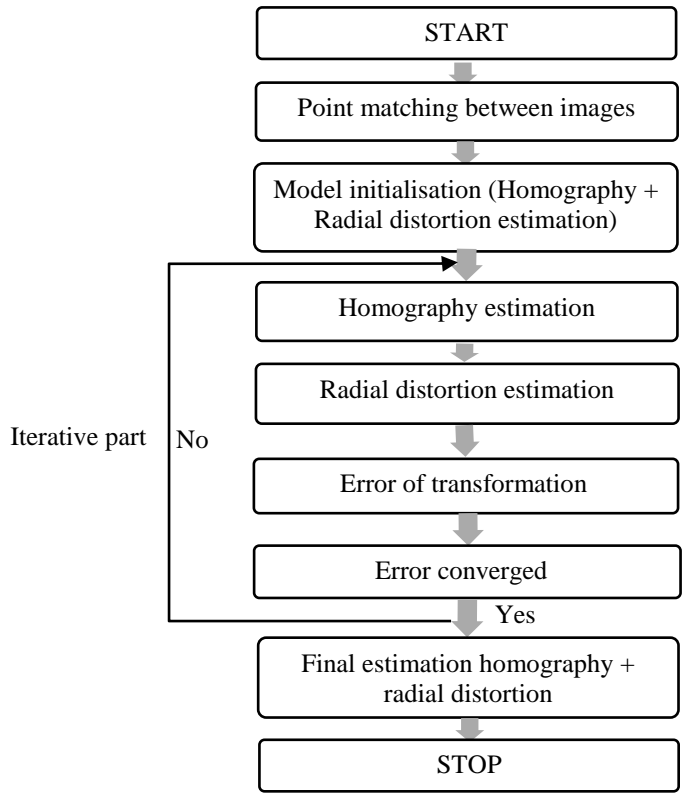

Figure 6. Flowchart of the superimposition method based on a homography and a radial distortion process.

The estimation process is as follows (Figure 6). The homography and the radial distortions are initialised using linear estimators. Then, an iterative estimation is performed independently for the homography and the radial distortions until the deformation error has converged. In order to ensure the error has converged, at each step a linear estimator is used as an initialiser of a non-linear (iterative) optimiser such as Levenberg-Marquardt (More, 1977; Bonnans *et al.*, 2006). Indeed, without initialisation, the non-linear estimator may converge towards a local minimum of the error (Hartley and Zisserman, 2004) (p. 110). The convergence criterion is defined by:

$$convergence = [err_n < \varepsilon] \text{ and } \left[\frac{err_n - err_{n-1}}{err_{n-1}} < tol\right] \text{ and } \left[\tilde{k} \text{ or } \{\tilde{k}_1 \text{and } \tilde{k}_2\} \in [-0.2 \,; 0.2]\right]$$

$$\text{and } [n < MaxIter]$$

$err$ is the mean error at iteration $n$, $\varepsilon = 0.01$ is the tolerance on the error, $tol = 0.01$ is a tolerance on the relative error between iterations and $MaxIter = 100$ is the maximum number of iterations. For each iteration, the normalised radial distortion parameters $\tilde{k}$ must be in the interval $[-0.2 \,;\, 0.2]$. If not, the algorithm stops and the model estimate with the smallest error is selected.

The final optimisation is performed jointly on the homography and the radial distortion with a linear estimator followed by a non-linear optimiser such as Levenberg-Marquardt (More, 1977; Bonnans *et al.*, 2006) for a single distortion or for two distortions the "trust region method" (Moré, 1983; Bonnans *et al.*, 2006) with the bounds $[-0.2 \,;\, 0.2]$ for the radial distortion parameters.

Notice: Even, if linear equations are introduced to estimate the model parameters, a final non-linear estimator is used for two reasons. Firstly, an estimate of the model is performed using the (non-linear) parametrisation of the matrix $A$ into a set of two angles and two scaling factors (Hartley and Zisserman, 2004) (p. 40), which have a direct physical interpretation. Secondly, the location of the correspondence points may be re-estimated jointly with the model parameters using the "Sampson error" or the "gold standard error" (Hartley and Zisserman, 2004) (p. 114). In this paper, the position of the correspondence points has not been re-estimated. It will be studied in a future paper.

### 2.5.1 Estimation of the model with two radial distortions.

When a different camera is used for each image of a pair, two radial distortions must be estimated, $k_1$ and $k_2$. Let us define linear estimators to initialise the non-linear estimators of the model parameters.

#### 2.5.1.1 Linear estimator of the radial distortion parameters $k_1$ and $k_2$

Equations (1) and (3) imply that:

$$k_1 k_2 \left[ \left(r_1^d r_2^d\right)^2 . D \right] + k_1 \left[ r_1^{d\,2} . D + r_1^{d\,2} . \bar{P}_2^d \right] + k_2 \left[ r_2^{d\,2} . D - r_2^{d\,2} . A \bar{P}_1^d \right] = -\left[ \bar{P}_2^d + D - A \bar{P}_1^d \right] \qquad (4)$$

with $D = C_2 - A C_1 - T$. Equation (4) is a quadratic equation with its unknowns $k_1$ and $k_2$: $k_1 k_2 [M_{12}] + k_1 [M_1] + k_2 [M_2] + [M_0] = 0$. The variables $M_i$, $i = \{0,1,2,12\}$, have two components $M_i = \left[ M_i^x, M_i^y \right]^t$, $M_i^x, M_i^y \in \mathbb{R}^N$. For each component, the equation is verified for each of the $N$ pairs of matched points. This leads to a set of two systems of quadratic equations. The set of equations corresponds to the intersection of two conics which as a unique pair of solutions according to Bernstein's theorem for bivariate polynomials (Sturmfels) (p. 37). This result can be explained in a more intuitive way; between a pair of matched points, there is one pair of solutions per equation (i.e. two pairs). As the solution must be valid for both systems of equations, there is only a unique solution pair $(k_1, k_2)$ valid for all the matched points. The other solutions are only valid for a single pair of matched points. The unique solution pair $(k_1, k_2)$ satisfies both systems of equations together and can be obtained by elimination of the second order term:

$$\left. \begin{array}{l} k_1 k_2 + \dfrac{M_1^x}{M_{12}^x} k_1 + \dfrac{M_2^x}{M_{12}^x} k_2 + \dfrac{M_0^x}{M_{12}^x} = 0 \\[2mm] k_1 k_2 + \dfrac{M_1^y}{M_{12}^y} k_1 + \dfrac{M_2^y}{M_{12}^y} k_2 + \dfrac{M_0^y}{M_{12}^y} = 0 \end{array} \right\} \Rightarrow k_1 \left[ \dfrac{M_1^x}{M_{12}^x} - \dfrac{M_1^y}{M_{12}^y} \right] + k_2 \left[ \dfrac{M_2^x}{M_{12}^x} - \dfrac{M_2^y}{M_{12}^y} \right] = - \left[ \dfrac{M_0^x}{M_{12}^x} - \dfrac{M_0^y}{M_{12}^y} \right]. \qquad (5)$$

The unique solution pair $(k_1, k_2)$ of the equation (5) is determined using the method of least squares. When the camera is the same for both images, only one radial distortion, $k = k_1 = k_2$, needs to be estimated in the equation (5).

#### 2.5.1.2 Linear estimator of the radial distortion parameters $k_1$ and $k_2$ and the homography

From equations (1) and (3), we obtain:

$$k_1 k_2 D \left[ \left(r_1^d r_2^d\right)^2 \right] + k_1 D \left[ r_1^{d\,2} \right] + k_1 \left[ r_1^{d\,2} . \bar{P}_2^d \right] + k_2 D \left[ r_2^{d\,2} \right] - k_2 A \left[ r_2^{d\,2} . \bar{P}_1^d \right] - A \left[ \bar{P}_1^d \right] + D[1] = -\left[ \bar{P}_2^d \right] \qquad (6)$$

Equation (6) is a linear equation with its variables in the bracket. It is valid for each component ($x$ and y) of the vectors $\bar{P}_1^d = \left[ \bar{P}_1^{d,x}, \bar{P}_1^{d,y} \right]^t$, $\bar{P}_2^d = \left[ \bar{P}_2^{d,x}, \bar{P}_2^{d,y} \right]^t$ and $D = [D^x, D^y]^t$, leading to a system of two equations:

$$\begin{cases} k_1 k_2 D^x \left[\left(r_1^d r_2^d\right)^2\right] + k_1 D^x \left[r_1^{d\,2}\right] + k_1 \left[r_1^{d\,2} \bar{P}_2^{d,x}\right] + k_2 D^x \left[r_2^{d\,2}\right] - k_2 a_{11} \left[r_2^{d\,2} \bar{P}_1^{d,x}\right] - k_2 a_{12} \left[r_2^{d\,2} \bar{P}_1^{d,y}\right] \\ \qquad\qquad -a_{11}\left[\bar{P}_1^{d,x}\right] - a_{12}\left[\bar{P}_1^{d,y}\right] + D^x[1] = -\left[\bar{P}_1^{d,x}\right] \\ k_1 k_2 D^y \left[\left(r_1^d r_2^d\right)^2\right] + k_1 D^y \left[r_1^{d\,2}\right] + k_1 \left[r_1^{d\,2} \bar{P}_2^{d,y}\right] + k_2 D^y \left[r_2^{d\,2}\right] - k_2 a_{21} \left[r_2^{d\,2} \bar{P}_1^{d,x}\right] - k_2 a_{22} \left[r_2^{d\,2} \bar{P}_1^{d,y}\right] \\ \qquad\qquad -a_{21}\left[\bar{P}_1^{d,x}\right] - a_{22}\left[\bar{P}_1^{d,y}\right] + D^y[1] = -\left[\bar{P}_2^{d,y}\right] \end{cases} \tag{7}$$

The 20 variables - in the brackets - correspond to the data. The 18 parameters of the set of both equations are estimated using the method of least squares applied to the set of both equations. As there are multiple relations between the 18 estimator parameters, only the 8 parameters of the model - $k_1, k_2$ and $H$ (through the intermediary of $A$ and $D$) - are unknown. An admissible solution for $k_1, k_2$ and $H$ is deduced from the 18 parameters of the linear estimator. The value of $k_1$ is directly obtained (3rd term of the 1st equation) as well as those of $D^x$ and $D^y$ (9th terms of both equations) and $a_{ij}$ (7th and 8th terms of both equations). The value of $k_2$ is deduced, e.g. from the 4th term of the 1st equation. Then, this admissible solution is introduced as the initialisation of the non-linear estimation.

### 2.6 Image warping

In order to analyse a large database, a fast algorithm of image warping is needed. Forward warping is time consuming and so we therefore use inverse warping. However, the registration model needs to be invertible (Wolberg, 1990). The radial distortion is modelled in equation (3) by a *division model* (Fitzgibbon, 2001). Wonpil (2003) and Park *et al.* (2009) have computed an approximate transformation for a standard distortion method. Here, the exact inversion of the *division model* is computed.

Given $r^u = \|P^u - C\| = \|\bar{P}^u\| \in \mathbb{R}$, the distance of the undistorted coordinates $P^u$ from the optic centre $C$, using equation (2), gives:

$$r^d = \left(1 + k\left(r^d\right)^2\right) r^u \tag{8}$$

Equations (2) and (8), implies that:

$$\bar{P}^u = \frac{r^u}{r^d} \bar{P}^d = W(\bar{P}^d) \tag{9}$$

In order to use invert warping, it is necessary to determine $W^{-1}$ transforming the undistorted points $P^u$ into the distorted points $P^d$. From equation (9), it is equivalent to determine $r^u$ knowing $r^d$.

Equation (8) is equivalent to: $kr^u r^{d\,2} - r^d + r^u = 0$, a second order equation in $r^d$. In the demonstration hereinafter, we show that its discriminant, $\Delta = 1 - 4kr^{u\,2}$, is strictly positive and that the inverse transformation $W^{-1}$ corresponds to the root:

$$r^d = W^{-1}(r^u) = \frac{1 - \sqrt{1 - 4kr^{u\,2}}}{2kr^u} \tag{10}$$

Demonstration: let us demonstrate (i) the positivity of the discriminant and (ii) the choice of the selected root.

We have: $\left.\begin{array}{l} r^u = \|P^u - C\| \leq \|C\| \\ |k| = \frac{|\bar{k}|}{1 + \|C\|^2} \leq \frac{2.10^{-1}}{1 + \|C\|^2} \end{array}\right\} \Rightarrow |k|r^u = \frac{|\bar{k}|}{1 + \|C\|^2} \|P^u - C\| \leq 2.10^{-1} \frac{\|C\|}{1 + \|C\|^2} \leq \frac{2.10^{-1}}{\|C\|}$

which implies that $\frac{1}{2|k|r^u} \geq \frac{\|C\|}{4.10^{-1}}$ and $4|k|r^{u\,2} \leq 8.10^{-1} < 1$.

Therefore, we have: $\Delta = 1 - 4kr^{u\,2} > 0$ (i) q.e.d.

When the discriminant is positive, there exists two roots: a) $\frac{1 + \sqrt{\Delta}}{2kr^u} \geq \frac{\|C\|}{4.10^{-1}} = 2.5\|C\|$, which corresponds to a solution where the pixel locations are rejected outside a given radius, $2.5\|C\|$, This is an aberrant solution. b) $\frac{1 - \sqrt{\Delta}}{2kr^u} \leq \frac{\|C\|}{4.10^{-1}} = 2.5\|C\|$, which corresponds to a solution where the pixel locations are inside a given radius, $2.5\|C\|$. This is the correct

solution. (ii) q.e.d.

Therefore, the transformation used is invertible (Equation (10)). An invertible image warping method compared to a non-invertible method dramatically reduces the computation time (Wolberg, 1990). We experimented a reduction from about 10 minutes to a few seconds (about 5 seconds) on a standard computer using Matlab (16GB RAM, processor Intel i7-4702HQ, 2.20GHz).

### 2.7    Experimental validation

### 2.7.1    Validation by a simulated montage

A simulated montage was created by superimposing two eye fundus images and deforming them according to the method presented by Lee *et al.* (2010). Real registered images were selected with an overlap percentage of 80% corresponding to the most common case in longitudinal databases. No modification of colour was done to the images. After adding equally spaced landmarks (i.e. the ground truth), the images were cut and deformed according to the model of Lee *et al.* (2010). An affine transformation has been used, with rotation scaling and shearing. Then, the image was modified by a projective distortion. The radius of the eye ball is equal to the ratio between the radius of the disk of the image divided by the observation angle of the camera (45 degrees). Then, the images were registered and the error was measured between the landmarks after registration and their true position (i.e the ground truth).

The quality of image superimposition was evaluated from the simulated montage for three approaches: (i) our method with a single radial distortion, (ii) our method with two radial distortions and for (iii) a state-of the art method using a quadratic model "gdbicp" (Yang *et al.*, 2007). The ground truth of the simulated montage allows a comparison between our model and the quadratic model "gdbicp". The duration of each superimposition method is also measured.

### 2.7.2    Validation with a public health database

Our method needs to be visually evaluated and compared to other state-of-the-art methods in public health database used for retinopathy screening. In order to assess the evolution of Diabetic Retinopathy several screening programs in the world are in existence. Trials have been performed in a database of 69 randomly selected patients coming from the Diabetic Eye Screening Wales (DESW) program in the United Kingdom (Thomas *et al.*, 2012). All patients had diabetes and different severity stages, including no signs, of retinopathy or maculopathy. All the selected patients have been screened annually for several years, seven years in average. For each of them we have kept two series of two examinations with an approximate screening interval of one year between the examination events. For each event exam, four images are available, two positions (nasal and macular) per eye. There were two series of images acquired for different years, with the first series are made up of 271 pairs of sufficient image quality (63% are acquired with different cameras) and the second series included of 268 pairs (9% are acquired with different cameras). In the first series, 11 pairs have a small overlapping area (about 30 % of the surface of the registered pair). For each position, we have performed the superimposition of the images between the two different examinations. All the retinal photographs were high quality images made according to a protocol including pupillary dilation.

In addition, in order to show that our method is not only useful on high quality images acquired following pupillary dilation but also on low quality images without dilation of the pupils prior to photography, a second validation test is performed to superimpose the images of low/poor quality for 5 patients with diabetes. The quality of image superimposition is checked for the 20 pairs of images (i.e. 4 per patient). The acquisitions conditions were significantly harsher compared to the high quality images and the quality of images was quite heterogeneous in part due to the lack of pupillary dilatation prior to photography. In this database, the patients with diabetes have two retinal examination with at least a year of interval using the same camera.

As there is no reference registration for the databases (i.e. no ground truth), for all pairs of images the superimposition has been visually checked by the same expert. The complete overlapping area of each pair has been carefully checked. The classification has been done in two categories: 1) no noticeable difference (i.e. correct) and 2) noticeable difference (i.e. incorrect). The category "incorrect" includes three sub categories: i) differences of the size of a small diameter vessel, ii) differences of the size of a large diameter vessel or iii) even larger. In this paper, the three subcategories have been grouped into a single category "incorrect". For each set of about 270 images, the duration of one check for one model is longer than an hour and a half. For each method and each pair of each set a first check has been done and a second check has been performed for verification by the same person. For almost all the pairs the results have been in accordance between both checks. In the rare cases of discrepancy, an adjudication has always been taken in the most restrictive way (i.e. "incorrect"). Therefore, the total duration of the check process was longer than thirty hours for the five methods tested hereinafter.

A visual evaluation has been performed on the two series of high quality images and on the series of low quality images in order to compare our approach to two other well established state-of-the-art whose code is publicly available: "gdbicp" (Stewart *et al.*, 2003; Yang *et al.*, 2007) or that we have reprogrammed (Lee *et al.*, 2007). In particular, we have tested five methods (see Table 3):

(i) our method on the normalised images with an automatic choice of one or two radial distortions according to the number of cameras used,

(ii) our model with a single radial distortion for all the original image pairs without colour stabilisation, in order to be similar to the state-of-the-art-method of (Lee *et al.*, 2007),

(iii) the state-of-the-art-method "gdbicp" with a quadratic model (Stewart *et al.*, 2003) on the original images,

(iv) "gdbicp" with a radial model on the original images (Yang *et al.*, 2007),

(v) and "gdbicp" with a quadratic model (Stewart *et al.*, 2003) on the normalised images.

Notice: To facilitate the comparison between (i) our model and (ii) the model similar to (Lee *et al.*, 2007), we have performed the model fitting with our method for both models, using the characteristics points extracted by SIFT, whereas (Lee *et al.*, 2007) use the centreline of the vessels and the branch centres for their model.

## 3 Results

### 3.1 Validation by a simulated montage

Using our method, with a single distortion, the mean registration error is of 0.86 pixels (standard deviation 1.75 pixels) in images of size 1568 x 2352 pixels (Figure 7) with vessels of maximum diameter of 30 pixels. The relative error respective to the image diagonal is 0.03%, and respective to the vessels is 2.9 %. With two distortions, the mean registration error is of 0.92 pixels (standard deviation 1.94 pixels) and the relative error is 0.03 % respective to the image and 3.1 % respective to the vessels (see Table 1). In the Figure 7, the error is mainly located on the external part of the superimposed pair of images.

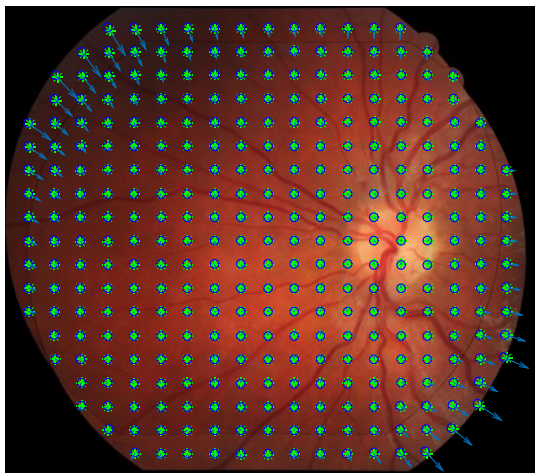

Figure 7. Validation of the superimposition model by registering a pair of images previously deformed. The green stars corresponds to the points of the reference image and the blue dots to the points of the current image. The arrows represent the registration errors between the two images.

Our method is compared to another state-of-the-art method "gdbicp" which superimposes eye fundus images with a quadratic model (Stewart *et al.*, 2003). Using the simulated montage, the mean registration error is of 2.44 pixels (standard deviation 1.64 pixels). As a reminder, the model of deformation proposed in this paper with a single radial distortion corresponds to the model of Lee *et al.* (2007), however, with a different method to estimate the parameters.

As a speed test, the duration of the different superimposition methods is measured with the images of the simulated montage (see Table 2). The experiments have been done on a recent computer (16GB RAM, processor Intel i7-4702HQ, 2.20GHz). The inverse warping method with one or two distortions lasts approximately 5.5 s using Matlab.

**Table 1.** Superimposition error for different methods of superimposition with the simulated montage

| Model | Mean (pixels) | Standard Deviation (pixels) | Mean relative error respective to the image (%) | Mean relative error respective to the vessels (%) |
|---|---|---|---|---|
| **Homography and 1 radial distortion** (normalised images) (our method) | **0.86** | **1.75** | **0.03 %** | **2.9 %** |
| **Homography and 2 radial distortions** (normalised images) (our method) | **0.92** | **1.94** | **0.03 %** | **3.1 %** |
| "gdbicp" quadratic (initial images) | 2.44 | 1.64 | 0.08 % | 8.1 % |

**Table 2.** Duration of each method of superimposition

| Model | Duration (seconds) |
|---|---|
| **Homography and 1 radial distortion** (normalised images) (our method) | 64.5 |
| **Homography and 2 radial distortions** (normalised images) (our method) | 71.7 |
| "gdbicp" quadratic (initial images) | 169 |

*3.2  Validation with a public health database and comparison to other state-of-the-art methods*

A visual evaluation is essential on the public health databases for the five methods previously described. Preliminary results have been presented in (Noyel *et al.*, 2017).

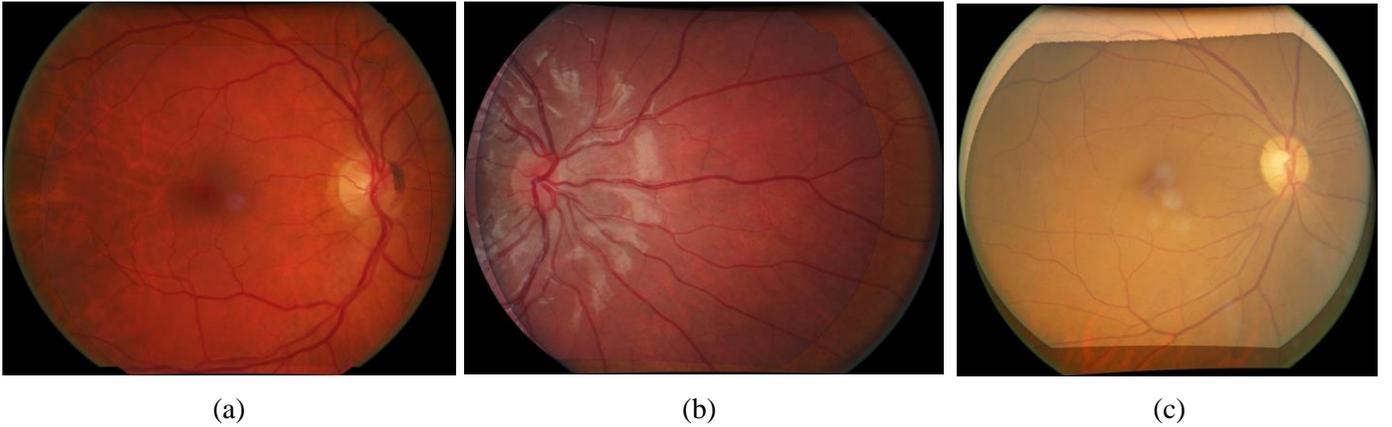

(a)   (b)   (c)

Figure 8. Superimposition of 3 pairs of eye fundus images with correction of one or two radial distortions. (a) Superimposition of the pair in the Figure 2 (a) and (b), with two radial distortions $\tilde{k}_1 = -0.0546$, $\tilde{k}_2 = -0.0646$. (b) Superimposition of the pair in the Figure 2 (c) and (d), with two radial distortions $\tilde{k}_1 = -0.0695$, $\tilde{k}_2 = -0.0536$. (c) Superimposition of the pair in the Figure 4 (a) and (b), with a single radial distortion $\tilde{k} = -0.0819$.

(i) First of all, our superimposition method is applied on standardised images. If the same camera acquired both images of a pair, our model is used with a single radial distortion. If both cameras were used, the model with two radial distortions is applied. In Figure 1 and in Figure 8, four examples of superimpositions with the radial distortion model are shown. One can notice the good quality of the superimposition. In the Figure 8, (a) and (b), two radial distortions have been used to correct the deformation of two different cameras whereas in the Figure 8 (c) a single radial distortion corrects the deformation of a single camera. The validation results for the complete database are presented in table 3. In the first series of 271 pairs, 2 pairs have small differences in the external part of the superimposition. These differences are of the size of the diameter of a vessel (Figure 10). When the percentage of overlapping surface is low (around 30 %) compared to the surface of the superimposed image, the superimpositions are correct in 92 % of the pairs and 96% if we consider the pairs in the same position. In the second series of 268 pairs of images acquired a few years later, the superimposition was successful for 98 % of the pairs. Within the pairs incorrectly superimposed, the external part was presenting differences whereas the central part was perfectly superimposed. For both series, the pairs with a sufficient overlap (greater than 50 %), no noticeable difference has been perceived between them.

**Table 3.** Percentage of correct superimpositions of image pairs of eye fundus in public health databases for different methods and models

| Model | 1st series - 271 pairs (63% different cameras) (high quality images) | 2nd series - 268 pairs (9% different cameras) (high quality images) | 3rd series - 20 pairs (similar cameras) (low quality images) |
|---|---|---|---|
| **(i) Homography and 1 or 2 radial distortions** (normalised images) (our method) | **92%** | **98%** | **100%** |
| (ii) Homography and 1 radial distortion (original images) (similar to (Lee *et al.*, 2007)) | 88% | 95% | 95% |
| (iii) "gdbicp" quadratic (original images) (Yang *et al.*, 2007) | 74% | 58% | 40% |
| (iv) "gdbicp" radial (original images) | 48% | 33% | 25% |
| (v) "gdbicp" quadratic (normalised images) | 88% | 87% | 100% |

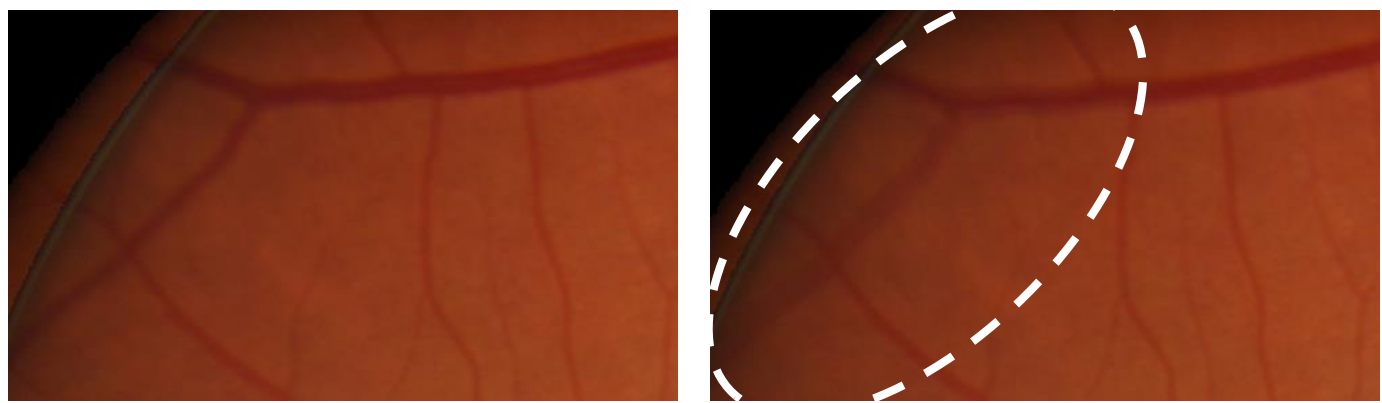

(a)                                                                     (b)

Figure 9. (a) Correct superimposition with our method based on a homography and two radial distortions with normalised images ($k_1 = -0.0616$ and $k_2 = -0.0685$). (b) Incorrect superimposition with noticeable differences (blurred vessel) using the method similar to (Lee *et al.*, 2007) on the initial images ($k = -0.0334$). The differences might be more visible in the electronic version.

(ii) Our approach is compared to the one similar to Lee *et al.* (2007) (Table 3). In the first series with 271 pairs (respectively second series with 268 pairs), the method similar to Lee *et al.* (2007) finds a superimposition solution for 88% of pairs (respectively 95%) while ours gives a correct result for 92% of the pairs (respectively 98%). In figure 9, with their approach, a noticeable difference appears near a vessel located on the external part of the image, where the effect of the radial deformation is the most important, while there is no noticeable difference using our method.

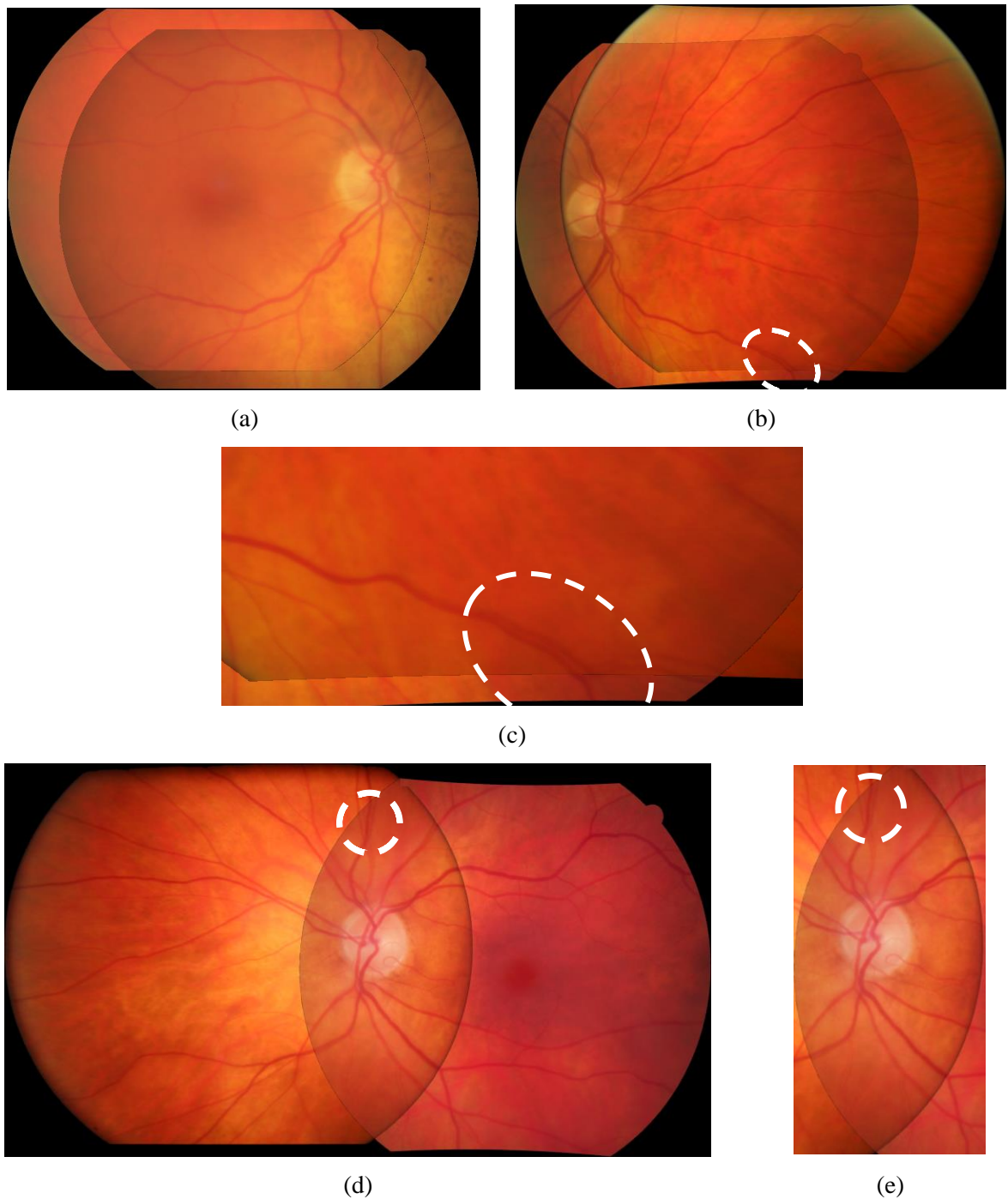

Figure 10. Examples of superimposition with our method. (a) Good superimposition. (b) Superimposition with small differences (in the white circle) and its enlargement (c). (d) Superimposition with small differences (in the white circle) and its enlargement (e). There is only a small overlap in the pair of (d) and (e).

(iii) Our method is compared to the state of the art method "gdbicp" (Stewart *et al.*, 2003; Yang *et al.*, 2007) which can make superimposition of images with a quadratic model (Table 3). In the first series with 271 pairs (respectively second series with 268 pairs), their method has found a superimposition solution for 74% of pairs (resp. 58%) while our method gives a correct result for 92% of the pairs (resp. 98%). (iv) In addition, when using with "gdbicp" a homography and two radial distortions using a multiplicative model - which is therefore not invertible - less pairs are registered (48% in the first series and 33% in the second series). The method of "gdbicp" with a quadratic model gives better results for eye fundus images than "gdbicp" with radial distortions because, "gdbicp" with a quadratic model was specifically designed to superimpose eye fundus images in Stewart *et al.* (2003).

In order to illustrate the robustness of our method, the superimposition is performed on the database of 5 patients with 20 pairs of low quality images. In each case, there was no noticeable difference in the superimposition (i) with our method, even on the border of the images. (ii) The method similar to the one of Lee *et al.* (2007) gives a correct superimposition for 95% of the pairs, (iii) "gdbicp" with a quadratic model 40% and (iv) "gdbicp" with a homography and two radial distortions 25%. In this second validation, our method gives better result than the others.

## 4 Interpretation & discussion

The accuracy of the superimposition methods has been validated on a simulated montage (Table 1). The superimposition error is smaller for our model including one or two radial distortions than with a quadratic model of superimposition such as the one used in "gdbcip" quadratic (Stewart *et al.*, 2003; Yang *et al.*, 2007; Adal *et al.*, 2014; Jian *et al.*, 2010). This confirms the results of Lee *et al.* (2010) who previously showed that quadratic models have a greater error than a homography and a radial distortion model. Such results demonstrate that our method gives a superimposition without noticeable difference, by a human observer. Therefore, this approach is suitable to perform an analysis in a large public health database.

In terms of speed, our method is 2.4 times faster than "gdbicp" quadratic (Table 2), even using a slower computer language (Matlab versus a compiled language). The duration results show that our method is computationally efficient to analyse large databases. The good speed results are due to the linear estimators of the model parameters and the use of an invertible model of deformation. In future, we will look for an even more efficient implementation using a compiled language, which is faster.

In order to assess the efficiency for public health purposes, our superimposition method has been validated on public health databases with high quality images of 69 patients (two series including 271 pairs and 268 different pairs) and 5 patients (20 pairs) with low quality images. In the first series, the majority of the pairs were acquired with two cameras while in the second series the majority were acquired with the same camera. Whatever is the number of cameras, there is no noticeable difference in the superimposed images if the overlap is sufficient (more than 50% about) (Table 3-i). For the pairs with a small overlapping area, we have developed another algorithm using, in addition to the matched points, the distance between the retinal vessels. This algorithm is similar to those described by Can *et al.* (2002) and Lee *et al.* (2010). These findings will be presented in a future paper. However, the interest of the superimposition is to compare the evolutions in public health databases over many years, which is only useful when the image overlap is large enough. This means that our method, efficient when the image overlap is sufficient, is suitable to be applied to analyse large databases.

(v) Finally, in order to show the influence of the colour stabilisation on the superimposition, the "gdbicp" method with a quadratic model is applied on images with colour stabilisation (i.e. normalised). The results obtained with colour stabilisation are much better than without: in the first series with 271 pairs, 88% vs. 48%, in the second series with 268 pairs, 87% vs. 33% and in the third series with 20 low quality pairs of 100 % vs. 25%.

Compared to the other state-of-the-art method of Lee *et al.* (2007) made of a homography and a single distortion (ii), for all the three series of images, our method (i) systematically improves the obtained results (Table 3). This result is due to the use of two radial distortions, instead of one, when images of different size are acquired by different cameras. This shows the importance of using an adapted model to the type of cameras used to acquire a pair of images. Therefore, our model goes further than the previous one (Lee *et al.*, 2007) that was not taking into account the change of camera. We underline that in (Lee *et al.*, 2007; Lee *et al.*, 2010) the authors were interested by large mosaicking with images acquired by the same camera during the same patient examination and therefore, they were using a corresponding database (i.e. images of similar colour and same distortion parameter). Our database is made of images of the patients acquires with at least a year of interval by different cameras and they may be of low quality.

Compared to the method "gdbicp" quadratic (Yang *et al.*, 2007) (iii), which uses a quadratic model, our method (i) better registers the eye fundus images. Our method still better superimposes the images than the method "gdbicp" quadratic with normalised images (v). However, when the image overlapping area is small (i.e. less than 50 %), "gdbicp" quadratic (v) better registers than our method. This is due to the detection of vessels landmarks, which makes the model fitting more efficient with a small overlapping area. However, in this image configuration, the superimposition solution given by "gdbicp" quadratic (iii, v) often presents a strong shearing deformation that has no physical justification. This means that the quadratic model is not well adapted when the overlapping area is small. This confirms the results obtained by Lee *et al.* (2010) who have shown that quadratic models introduces more errors than models based on a homography and a radial distortion. Indeed, with 12 parameters to be fitted (vs 8 parameters for our model – 6 for the homography and 2 for the radial distortions) quadratic models have more degree of freedom that is a source of additional errors. Therefore, our model goes further than the previous approaches with quadratic models (Can *et al.*, 2002; Lee *et al.*, 2010; Jian *et al.*, 2010; Adal *et al.*, 2014). When the overlapping area is small (less than 50 %), the method "gdbicp" quadratic with normalised images (v) is complementary to our method (i) due to the detection of vessel landmarks.

To summarise the comparison, our method (i) better superimposes the images of our public health databases (Table 3). The second model is the one of (Lee *et al.*, 2007) (ii), whose results are systematically improved by our method. The third model is "gdbicp" quadratic (Yang *et al.*, 2007) (iii) which is improved when normalising the images (v). The method gdbicp" quadratic, and also the method described in (Lee *et al.*, 2007), is based on the detection of vessel landmarks. Therefore, when they are used on images with a small overlapping, the vessel detection makes them more efficient and complementary to our method. We will present such improvement for our method in a future paper.

The obtained results with the validation montage and with the public health databases demonstrate that our model better corrects the errors coming from the different positions of the patient during image acquisition, the change in the camera employed (resolution and optical lenses) and the projection of a 3D scene onto a plane and the variability of colour between images.

In addition, the colour stabilisation step is useful when images have strong contrast variations. The results obtained with "gdbicp" quadratic with the colour stabilisation (v), which are better than and without (iii), show the importance of using normalised images for colour in public health databases (Table 3). In our method (i), the SIFT points are extracted and matched on the normalised images for colour contrast before estimating the registration model. Experiments have shown that using normalised images gives more robust results when extracting and matching the SIFT points, even if these points are known to be robust to colour variations (Lowe, 2004; Vedaldi and Fulkerson, 2008). Importantly, even when using normalised images with "gdbicp", our method remains better.

Moreover, our method has been tested for images acquired with a field of view (FOV) of 45° which are used for DR screening by single-field fundus photography (Williams *et al.*, 2004). In addition, our approach could be useful for automatic detection of referral patients due to Diabetic Retinopathy (Fleming *et al.*, 2010; Decencière *et al.*, 2013; Abramoff *et al.*, 2013; Quellec *et al.*, 2016).

Despite the good results obtained, our method does have a limitation. For images with a smaller overlap (e.g. 30% of the surface of the mosaic image), the superimposition may present small differences on the external part. To address this issue, we have developed another algorithm using in addition to the matched points, the distance between the vessels. It will be presented in a future paper.

## 5 Conclusions

We have therefore successfully achieved a new method to answer the main challenges to superimpose eye fundus images coming from large public health databases. In addition to the previously existing methods, ours has been designed to deal with changes in terms of camera, lens, image resolution and strong colour variations between two retinal exams. The presented method consists of fitting a registration model composed of a homography and one or two radial distortions on salient points extracted in images after colour stabilisation. The choice between the number of radial distortions is made automatically depending of the number of cameras used to acquire the pair of images. Then, the image warping is performed using a *division model,* which is invertible, which makes it fast to compute. Linear equations are introduced to estimate the parameters of the model in a fast way, followed by non-linear estimators. Importantly, the method is easy to use and does not require to extract intrinsic characteristics such as the vessels or their branch points.

The efficiency of our method has been validated on a simulated montage with a superimposition error, which is less than the error of a quadratic model. Its efficiency has also been demonstrated on public health databases with high quality images and images of lower quality, due to differences in the conditions of acquisition, and pairs of images acquired with the same or different cameras. Nevertheless, the results show that there is no noticeable difference between the images from two examinations with the eye in the same position (nasal or macular). The superimposition is correct in more than 92% of the cases (96% with an overlapping area greater than 50 %). In these databases, our method better superimposes the images than the two state-of-the-art methods of (Lee *et al.*, 2007) and of "gdbicp" (Stewart *et al.*, 2003; Yang *et al.*, 2007).

**Conflict of interest**

No conflict of interest.

**Funding**

This research did not receive any specific grant from funding agencies in the public, commercial, or not-for-profit sectors.